\documentclass[11pt,a4paper]{article}
\usepackage[hyperref]{emnlp-ijcnlp-2019}
\usepackage{times}
\usepackage{latexsym}
\usepackage{booktabs}
\usepackage{graphicx}
\usepackage{xcolor}
\usepackage{url}
\usepackage{xcolor}
\usepackage{multirow}
\usepackage{comment}
\usepackage{tabularx}
\usepackage{colortbl}
\usepackage{tablefootnote}
\usepackage{threeparttable}
\usepackage{hhline}
\usepackage{enumitem}

\usepackage{amsmath,amsfonts,amssymb,bm} 

\aclfinalcopy %

\newcommand{\eat}[1]{\ignorespaces}

\title{Giving BERT a Calculator: Finding Operations and Arguments with Reading Comprehension}

\author{Daniel Andor, Luheng He, Kenton Lee, Emily Pitler \\
    Google Research \\
    {\tt \{andor, luheng, kentonl, epitler\}@google.com } \\}

\date{}

\begin{document}
\maketitle
\begin{abstract}
Reading comprehension models have been successfully applied to extractive text answers, but it is unclear how best to generalize these models to abstractive numerical answers.
We enable a BERT-based reading comprehension model to perform lightweight numerical reasoning.
We augment the model with a predefined set of executable `programs' which encompass simple arithmetic as well as extraction.
Rather than having to learn to manipulate numbers directly, the model can pick a program and execute it.
On the recent Discrete Reasoning Over Passages (DROP) dataset, designed to challenge reading comprehension models, 
we show a  33\% absolute improvement by adding shallow programs.
The model can learn to predict new operations when appropriate in a math word problem setting (Roy and
Roth, 2015) with very few training examples.
\end{abstract}

\section{Introduction}

End-to-end reading comprehension models have been increasingly successful at extractive question answering. 
For example, performance on the SQuAD 2.0 \cite{squad2} benchmark has improved from 66.3 F1 to 89.5\footnote{\url{https://rajpurkar.github.io/SQuAD-explorer/}} in a single year.
However, the Discrete Reasoning Over Passages (DROP) \cite{DROP} dataset demonstrates that as long as there is quantitative reasoning involved, there are plenty of relatively straightforward questions that current extractive QA systems find difficult to answer.
Other recent work has shown that even state-of-the-art neural models struggle with numerical operations and quantitative reasoning when trained in an end-to-end manner \cite{Saxton2019AnalysingMR,Ravichander2019EQUATEAB}.
In other words, even BERT \cite{BERT} is not very good at doing simple calculations.

\begin{table}[t]

\newcolumntype{Y}{>{\arraybackslash}X}
\small
\centering

\begin{tabularx}{\linewidth}{Y}
\toprule

\emph{How many more Chinese nationals are there than European nationals?} \\

\midrule

The city of Bangkok has a population of 8,280,925 ...the census showed that it is home to 81,570 Japanese and \textcolor{purple}{\textbf{55,893}} Chinese nationals, as well as 117,071 expatriates from other Asian countries, \textcolor{teal}{\textbf{48,341}} from Europe, 23,418 from the Americas,...\\

\midrule

{\bf NAQANet}: \textcolor{purple}{$-$55893}\\

{\bf Ours}: $\texttt{Diff}$(\textcolor{purple}{55893}, \textcolor{teal}{48341}) = {\bf 7552}\\

\bottomrule

\end{tabularx}
\caption[Caption for DROP example]{
    Example from the DROP %
    development set.  The correct answer %
    is not explicitly stated in the passage and instead must be computed.
    The NAQANet model\footnotemark \cite{DROP} predicts a negative number of people, whereas our model predicts that an operation $\texttt{Diff}$ should be taken and identifies the two arguments.
}
\label{tab:example}
\end{table}
\footnotetext{\url{https://demo.allennlp.org/reading-comprehension/NzQwNjg1}}

In this work, we extend an extractive QA system with numerical reasoning abilities. 
We do so by asking the neural network to synthesize small programs that can be executed.
The model picks among simple programs of the form $\texttt{Operation}(\text{args}, ...)$, where the possible operations include span extraction, answering yes or no,
and arithmetic. %
For math operations, the arguments are pointers to numbers in the text and, in the case of composition, other operations. 
In this way, the burden of actually doing the computation is offloaded from the neural network to a calculator tool.
The program additionally provides a thin layer of interpretability that mirrors some of the reasoning required for the answer.
For example, in Table~\ref{tab:example},
the model predicts subtraction ($\texttt{Diff}$) over two numbers in the passage, and executes it to produce the final answer.

We start with a simple extractive question answering model based on BERT \cite{BERT}, and show the following:
\begin{enumerate}[topsep=0pt,itemsep=0ex,partopsep=1ex,parsep=1ex]

\item Predicting unary and binary math operations with arguments resulted in significant improvements on the DROP dataset.

\item
Our model can smoothly handle more traditional reading comprehension inputs as well as math problems with new operations. 
Co-training with the CoQA  \cite{COQA} dataset %
improved performance on DROP. 
The DROP+CoQA trained model had never seen multiplication or division examples, but %
can learn to predict these two ops when appropriate in a math word problem setting \cite{Roy2015SolvingGA} with very few training examples.

\end{enumerate}

\section{Background and Related Work}

\paragraph{Discrete Reasoning over Paragraphs (DROP)} \cite{DROP} is a reading comprehension task that requires discrete reasoning. 
Inspired by semantic parsing tasks where models need to produce executable `programs', it keeps the open-domain nature of reading comprehension tasks such as SQuAD 2.0 \cite{squad2}.
As shown in Table~\ref{tab:example}, the system needs to perform fuzzy matching between \textit{``from Europe''} and \textit{``European nationals''} in order to identify the arguments.

\paragraph{Numerically-aware QANet (NAQANet)} \cite{DROP} is the current state-of-the-art\footnote{\url{https://leaderboard.allenai.org/drop/submissions/public}} system for DROP. 
It extends the QANet model \cite{qanet} with predictions for numbers (0--9) and summation operations. 
For the latter, it performs a 3-way classification (plus, minus, and zero) on all the numbers in the passage. %

While certain binary operations are expressible efficiently with flat sign prediction, it is difficult to generalize the architecture.
Moreover, each number is tagged independently, which can cause global inconsistencies;
for instance, in Table~\ref{tab:example} it assigns a single minus label and no plus labels, leading to a prediction of negative people.

\paragraph{Mathematical Word Problems} have been addressed with a wide variety of datasets and approaches; see \newcite{Zhang2018TheGO} for an overview.  
One such dataset of arithmetic problems is the Illinois dataset \cite{Roy2015SolvingGA}. 
The problems are posed in simple natural language that has a specific, narrow domain,
For example: \emph{``If there are 7 bottle caps in a box and Linda puts 7 more bottle caps inside, how many bottle caps are in the box?''}.
Unlike DROP, the problems are typically 1--3 sentences long and do not require reading complex passages.
Instead, the main challenge is mathematical reasoning. 
According to \newcite{Zhang2018TheGO}, the current state of the art uses syntactic parses and deterministic rules to convert the input to logical forms \cite{Liang2016ATS}.

\begin{table*}[]
\centering\small
\newcolumntype{Y}{>{\arraybackslash}X}
\newcolumntype{L}{>{\hsize=.32\hsize}Y}
\newcolumntype{C}{>{\hsize=.48\hsize}Y}
\newcolumntype{R}{>{\hsize=.2\hsize}Y}
\newcommand{\colindent}{\;}
\setlength{\tabcolsep}{.5em}
\begin{tabularx}{\linewidth}{l LCR}
\toprule

 & \textbf{Derivations} & \textbf{Example Question} & \textbf{Answer Derivation} \\
\midrule
\emph{Literals} &
{\sc Yes}, {\sc No}, {\sc Unknown}, 0, 1 ..., 9 & How many field goals did Stover kick? &  4 \\
\cmidrule(lr){1-4}

\emph{Numerical} &
$\texttt{Diff100}: n_0 \rightarrow 100 - n_1$ & How many percent of the national population does not live in Bangkok? & $100 - 12.6 = 87.4$ \\

& $\texttt{Sum}: n_0, n_1 \rightarrow n_0 + n_1\;\;\;\;$ \;\;\;\;
\; as well as: $\texttt{Diff}$, $\texttt{Mul}$, $\texttt{Div}$ 
& How many from the census were in Ungheni and Cahul? & $32,828 + 28,763 = 61591$\\ 

\cmidrule(lr){1-4}

\emph{Text spans} &
$\texttt{Span}: i,j \rightarrow s$ & Does Bangkok have more Japanese or Chinese nationals? & ``Japanese'' \\

\cmidrule(lr){1-4}
\emph{Compositions} &
$\texttt{Merge}: s_0, s_1 \rightarrow \{ s_0, s_1\}$ &
 What languages are spoken by more than 1\%, but fewer than 2\% of Richmond's residents? & ``Hmong-Mien languages'', ``Laotian''   \\

& $\texttt{Sum3}: n_0, n_1, n_2\rightarrow (n_0 + n_1) + n_2$ 
& How many residents, in terms of percentage, speak either English, Spanish, or Tagalog? &
$\texttt{Sum}(64.56,23.13)+2.11 =89.8$ \\

\bottomrule

\end{tabularx}

\caption{Operations supported by the model. $s, n$ refer to arguments of type \emph{span}  and \emph{number}, respectively. $i, j$ are the start and end indices of span $s$. The omitted definitions of \texttt{Diff}, \texttt{Mul}, and \texttt{Div} are analogous to \texttt{Sum}.}
\label{tab:operations}

\end{table*}

\section{Model}
We extend a BERT-based extractive reading comprehension model with a lightweight extraction and composition layer. 
For %
details of the BERT architecture see \newcite{BERT}. %
We only rely on the representation of individual tokens that are jointly conditioned on the given question $Q$ and passage $P$. 
Our model predicts an answer by selecting the top-scoring \emph{derivation} (i.e.~program) and executing it.

\paragraph{Derivations}

We define the space of possible derivations $\mathcal{D}$ as follows:
\begin{itemize}[topsep=1pt,itemsep=-1ex,partopsep=1ex,parsep=1ex]
    \item \emph{Literals}: $\{ \textsc{Yes}, \textsc{No}, \textsc{Unknown}, 0,\ldots9\}$.
    \item \emph{Numerical operations}: including various types of numerical compositions of numbers\footnote{Numbers are heuristically extracted from the text.}, such as \texttt{Sum} or \texttt{Diff}. 
    \item \emph{Text spans}: composition of tokens into text spans up to a pre-specified length.
    \item \emph{Composition of compositions}: %
    we only consider two-step compositions, including merging text spans and nested summations.
\end{itemize}

The full set of operations are listed in Table~\ref{tab:operations}. 
For example, \texttt{Sum} is a numerical operation that adds two numbers and produces a new number.
While we could recursively search for compositions with deep derivations, here we are guided by what is required in the DROP data and simplify inference by heavily restricting multi-step composition. 
Specifically, spans can be composed into a pair of merged spans (\texttt{Merge}), and the sum of two numbers (\texttt{Sum}) can subsequently be summed with a third (\texttt{Sum3}).
The results in Table~\ref{tab:drop-results} show the dev set oracle performance using these shallow derivations, by answer type.

\paragraph{Representation and Scoring}
For each derivation $d\in \mathcal{D}$, we compute a vector representation $\mathbf{h}_d$ and a scalar score $\rho(d, P, Q)$ using the BERT output vectors. 
The scores $\rho$ are used for computing the probability $P(d\mid P, Q)$ as well as for pruning.
For brevity, we will drop the dependence on $P$ and $Q$ in this section.

\emph{Literals} are scored as
    $\rho(d) = \mathbf{w}_d^{\intercal} \textsc{MLP}_{\text{lit}} (\mathbf{h}_{\texttt{CLS}})$,
where $\mathbf{h}_{\texttt{CLS}}$ is the output vector at the $\texttt{[CLS]}$ token of the BERT model \cite{BERT}.

\emph{Numeric operations} use the vector representations $\mathbf{h}_i$ of the first token of each numeric argument.
Binary operations are represented as
\begin{align}
    \mathbf{h}_{d} &= \textsc{MLP}_{\text{binary}} (\mathbf{h}_i, \mathbf{h}_j, \mathbf{h}_i \circ \mathbf{h}_j) \label{eq:binaryrep} %
\end{align}
and scored as $\rho(d) = \mathbf{w}_{op}^{\intercal} \mathbf{h}_{d}$, 
where $\mathbf{h}_{d}$ represents the binary arguments and $op$ is the operation type. $\circ$ is the Hadamard product.
Unary operations such as $\texttt{Diff100}$ are scored as
$\mathbf{w}_{op}^{\intercal}\textsc{MLP}_{\text{unary}} (\mathbf{h}_i)$.

\emph{Text spans} are scored as if they were another binary operation taking as arguments the start and end indices $i$ and $j$ of the span~\cite{Lee2017LearningRS}:
\begin{align}
    \mathbf{h}_{d} &= \textsc{MLP}_{\text{span}} (\mathbf{h}_i, \mathbf{h}_j) \label{eq:spanrep} %
\end{align}
and scored as $\rho(d) = \mathbf{w}_{\text{span}}^{\intercal} \mathbf{h}_{d} \notag$.  

\emph{Compositions of compositions} are scored with the vector representations of its children. 
For example, the ternary $\texttt{Sum3}$, comprising a $\texttt{Sum}$ and a number, is scored with $\mathbf{w}_{\texttt{Sum3}}^{\intercal} \textsc{MLP}_{\texttt{Sum3}} (\mathbf{h}_{d0}, \mathbf{h}_k)$, where $\mathbf{h}_{d0}$ corresponds to the representation from the first $\texttt{Sum}$, and $\mathbf{h}_k$ is the representation of the third number. The \emph{composition} of two spans is scored as  $\mathbf{w}_{\texttt{Merge}}^{\intercal} \textsc{MLP}_{\texttt{Merge}} (\mathbf{h}_{d0}, \mathbf{h}_{d1}, \mathbf{h}_{d0} \! \circ \! \mathbf{h}_{d1})$, where $\mathbf{h}_{d0}$ and $\mathbf{h}_{d1}$ are span representations from \eqref{eq:spanrep}.
The intuition for including $\mathbf{h}_{d0}~\circ~\mathbf{h}_{d1}$ is that it encodes span similarity, and spans with similar types are more likely to be merged.

This strategy differs from the NAQANet baseline in a few ways.  
One straightforward difference is that we use BERT as the base encoder rather than QANet.
A more meaningful difference is that we model all derivations in the unified op scoring framework described above, which allows
generalizing to new operations, whereas NAQANet would require more large-scale changes to go beyond addition and subtraction.
Generalizing the model to new ops is a case of extending the derivations and scoring functions. 
In Section~\ref{sec:experiments}, we will show the impact of incrementally adding \texttt{Diff100}, \texttt{Sum3}, and \texttt{Merge}.

\subsection{Training} \label{sec:training}

We used exhaustive pre-computed oracle derivations $\mathcal{D}^*$ following \newcite{DROP}. 
We marginalized out all derivations $d^*$ that lead to the answer\footnote{In practice we capped the number of derivations at 64, which covers 98.7\% of the training examples.} and minimized:
\begin{align}
    \mathcal{J}(P, Q, \mathcal{D}^*) & = - \log \sum_{d^* \in \mathcal{D}^*} P(d^* \mid P, Q) \nonumber \\
    P(d \mid P, Q) & = \frac{\exp \rho(d, P, Q)}{\sum_{d'} \exp \rho(d', P, Q)} \nonumber
\end{align}
If no derivation lead to the gold answer ($\mathcal{D}^*$ is empty), we skipped the example.

\paragraph{Pruning}
During inference, the $\texttt{Merge}$ and $\texttt{Sum3}$ operations are composed from the results of $\texttt{Span}$ and $\texttt{Sum}$ operations, respectively.
The space of possible results of $\texttt{Merge}$ is quadratic in the number $|\mathcal{S}|$ of possible spans.
With $|\mathcal{S}| \sim 10^4$, the complete set of $\texttt{Merge}$ instances becomes overwhelming. 
Similarly, with $|\mathcal{N}| \sim 100$ numbers in each passage, there are millions of possible $\texttt{Sum3}$ derivations.
To do training and inference efficiently, we kept only the top 128 $\texttt{Span}$ and $\texttt{Sum}$ results when computing $\texttt{Merge}$ and $\texttt{Sum3}$.\footnote{During training, the pruned arguments had recall of 80--90\% after 1 epoch and plateaued at 95--98\%.}

\begin{table*}[th]
\newcolumntype{Y}{>{\arraybackslash}X}
\newcommand{\colindent}{\;}
\small
\centering

\begin{tabularx}{\linewidth}{@{} l c c c c c c c c c c c c c @{}} 
\toprule
  & \it Oracle
  & \multicolumn{2}{c}{Overall Dev} 
  & \multicolumn{2}{c}{Overall Test}
  & \multicolumn{2}{c}{Date (1.6\%)}
  & \multicolumn{2}{c}{Number (62\%)}
  & \multicolumn{2}{c}{Span (32\%)}
  & \multicolumn{2}{c}{Spans (4.4\%)} \\
  \cmidrule(lr){3-4} \cmidrule(lr){5-6} \cmidrule(lr){7-8} \cmidrule(lr){9-10} \cmidrule(lr){11-12} \cmidrule(lr){13-14}
  & Dev EM & EM & F1 & EM & F1
  & EM & F1 & EM & F1 & EM & F1 & EM & F1  \\
 \midrule
  NAQANet
  & & 46.75 & 50.39 & 44.24 & 47.77 & 32.0 & 39.6 & 44.9 & 45.0 & 58.2 & 64.8 & 0.0 & 27.3 \\
  \cmidrule(lr){1-14}
  Our basic\footnotemark & \it 80.03 & 
  66.50   &   69.91 & - & - & 
  57.0 & 65.1 & 65.8 & 66.1 & 78.0 & 82.6 & 0.0 & 35.7 \\
  \ +\texttt{Diff100} & \it 88.75 &
  75.52   &   78.82 & - & - & 
  53.6 & 61.3 & 80.3 & 80.5 & 78.4 & 82.8 & 0.0 & 35.8 \\
  \ +\texttt{Sum3} & \it 90.16 &
  76.70   &   80.06 & - & - & 
  58.0 & 64.6 & 81.9 & 82.1 & 78.9 & 83.4 & 0.0 & 36.0 \\
  \ +\texttt{Merge} & \it 93.01 &
  76.95   &   80.48 & - & - & 
  58.1 & 61.8 & 82.0 & 82.1 & 78.8 & 83.4 & 5.1 & 45.0 \\
  \ +CoQA & \it 93.01 & 
  \bf 78.09 & \bf 81.65 & \bf 76.96 & \bf 80.53 & 
  \bf 59.5 & \bf 66.4 & \bf 83.1 & \bf 83.3 & \bf 79.8 & \bf 84.3 & \bf 6.2 & \bf 47.0 \\
\cmidrule(lr){1-14}
  \ +Ensemble & \it 93.01 & 
  \bf 78.97 & \bf 82.56 & \bf 78.14 & \bf 81.78 & 
  \bf 59.7 & \bf 67.7 & \bf 83.9 & \bf 84.1 & \bf 81.1 & \bf 85.4 & 6.0 & 47.0 \\
\cmidrule(lr){1-14}
  \it Oracle & & \it 93.01 & & & &
  \it 71.6 && \it 94.5 && \it 95.8 && \it 60.5  \\
 \bottomrule
\end{tabularx}
\caption{
Accuracies on the DROP dev and test set in terms of exact match (EM) and token-level F1. 
The righthand columns show the performance breakdown with different answer types on the development set.  
The largest improvements come from \emph{Date}, \emph{Number}, and \emph{Spans} (answers with multiple spans). 
\emph{Oracle} rows and columns indicate the performance that could be achieved by perfect selections of derivations.
The ensemble used 6 models.
}
\label{tab:drop-results}
\end{table*}

\paragraph{Spurious ambiguities}
Of the answers for which we could find at least one oracle derivation, 36\% had two or more alternatives.
During training, the model became effective at resolving many of these ambiguities.
We monitored the entropy of $P(d^* \mid P, Q)$ for the ambiguous examples as training progressed.
At the start, the entropy was 2.5 bits, which matches the average ambiguous oracle length of $\sim 6$ alternatives.
By the end of 4 epochs, the average entropy had dropped to $<0.2$ bits, comparable to a typical certainty of 95--99\% that one of the derivations is the correct one.

\section{Experiments}
\label{sec:experiments}

Our main experiments pertain to DROP \cite{DROP},
using DROP and, optionally, CoQA \cite{COQA} data for training.
Pre-processing and hyperparameter details are given in the supplementary material.
In addition to full DROP results, 
we performed ablation experiments for the incremental addition of the \texttt{Diff100},
\texttt{Sum3}, and
\texttt{Merge} operations, and finally the CoQA training data.
We ran on the CoQA dev set, to show that the model co-trained on CoQA can still perform traditional reading comprehension.
To investigate our model's ability to do symbolic reasoning at the other extreme,
we performed few-shot learning experiments on the Illinois dataset of math problems \cite{Roy2015SolvingGA}.

\subsection{DROP Results}

As shown in Table~\ref{tab:drop-results}, our model achieves over 50\% relative improvement (over 33\% absolute) over the previous state-of-the-art NAQANet system.
The ablations indicate that the improvements due to the addition of extra ops (\texttt{Diff100}, \texttt{Sum3}, \texttt{Merge}) are roughly consistent with their proportion in the data.
Specifically, the \texttt{Diff100} and \texttt{Sum3} derivations increase the oracle performance by 8.7\% and 1.4\% respectively, corresponding to model improvements of roughly 9\% and 1.1\%, respectively.
Answers requiring two spans occur about 2.8\% of the time, which is a 60.4\% proportion of the {\em Spans} answer type.
\texttt{Merge} only improves the {\em Spans} answer type by 9\%, which we think is due to the significant 11:1 class imbalance between competing single and multiple spans.
As a result, multiple spans are under-predicted, leaving considerable headroom there.

Pre-training on CoQA then fine-tuning on DROP lead to our best results on DROP, reported in Table~\ref{tab:drop-results}.
After fine-tuning on DROP, the model forgot how to do CoQA, with an overall F1 score of 52.2 on the CoQA dev set. 
If one prefers a model competent in both types of input, then the forgetting can be prevented by fine-tuning on both CoQA and DROP datasets simultaneously.
This resulted in dev set F1 scores of 82.2 on CoQA and 81.1 on DROP.
The CoQA performance is decent and compares well with the pre-trained model performance of 82.5.
The 0.5\% drop in DROP performance is likely attributable to the difference between pre-training versus fine-tuning on CoQA.

We ensembled 6 models (3 seeds $\times$ 2 learning rates) for an additional 1\% improvement.

\footnotetext{The ``basic'' model includes all $\mathcal{D}_{\text{direct}}$, all $\mathcal{S}$, and the simple binary operations \texttt{Sum} and \texttt{Diff}.}

\begin{table}[t]
\small\centering

\begin{tabularx}{0.59\linewidth}{ ll l} 
\toprule
\newcite{Roy2015ReasoningAQ} & 73.9 \\
\newcite{Liang2016ATS} &\bf 80.1 \\
\newcite{Wang2018MathDQNSA} & 73.3 \\
\cmidrule(lr){1-2}
Our basic: IL data                     & 48.6 $\pm$ 5.3 \\
\ + \texttt{Mul} and \texttt{Div} & 74.0 $\pm$ 6.0 \\
\ + DROP data                   & \bf 83.2 $\pm$ 6.0 \\
\bottomrule
\end{tabularx}
\caption{Accuracy on the Illinois (IL) dataset\footnotemark of 562 single-step word problems, 
using the five cross-validation folds of \newcite{Roy2015SolvingGA}.
Standard deviations were computed from the five folds.
Roughly half the questions require the use of \texttt{Sum} and \texttt{Diff},
and half require \texttt{Mul} and \texttt{Div}.
}
\label{tab:il-results}
\end{table}

\subsection{Results on Math Word Problems}
\label{sec:mawps}

We trained our model on the Illinois math word problems dataset \cite{Roy2015SolvingGA}, 
which contains answers requiring multiplication and division---operations not present in DROP---as well as addition and subtraction, 
in roughly equal proportion.
Given the small ($N=562$) dataset size, training and evaluation is done with five-fold cross-validation on a standardized set of splits.
As shown in Table~\ref{tab:il-results}, when we added \texttt{Mul} and \texttt{Div} to our basic DROP operations,
the model was able to learn to use them.
Transferring from the DROP dataset further improved performance beyond that of \newcite{Liang2016ATS},
a model specific to math word problems that uses rules over dependency trees.
Compared to other more general systems, our model outperforms the deep reinforcement learning based approach of \newcite{Wang2018MathDQNSA}.

\footnotetext{\url{https://cogcomp.org/page/resource_view/98}}

\section{Conclusions and Future Work}

We proposed using BERT for reading comprehension combined with lightweight neural modules for computation in order to smoothly handle both traditional factoid question answering and questions requiring symbolic reasoning in a single unified model.
On the DROP dataset, which includes a mix of reading comprehension and numerical reasoning, our model achieves a 33\% absolute improvement over the previous best.
The same model can also do standard reading comprehension on CoQA, and focused numerical reasoning on math word problems.
We plan to generalize this model to more complex and compositional answers, with better searching and pruning strategies of the derivations.

\section*{Acknowledgements}

We would like to thank Chris Alberti, Livio Baldini Soares, and Yoon Kim for tremendously helpful discussions, 
and we are grateful to all members of the Google Research Language team.

\bibliography{emnlp-ijcnlp-2019}
\bibliographystyle{acl_natbib}

\end{document}


\maketitle
\appendix

\section{Pre-processing and Hyper-parameters}

After processing the input with the standard BERT tokenizer, we extracted the locations and values of numbers.
We allowed for up to 128 numbers per document, with typical documents having 10--20.
The maximum length of spans is set to 32. %
Documents longer than 512 tokens are split up.

We use a whole-word masked version of BERT similar to \newcite{Sun2019ERNIEER}. 
Unless otherwise indicated, we fine-tuned \texttt{BERT$_{\text{LARGE}}$} with a batch size of 32 over a small grid of hyperparameters: 
We varied only the learning rate in the range $2{\rm e}{-5}$ to $5{\rm e}{-5}$ and the number of epochs between 1--5.
We did random restarts with 2--4 random seeds.

\section{Examples of Wins and Losses}

We selected a few examples of the wins and losses of our model in Table \ref{tab:win1}, \ref{tab:win2}, and \ref{tab:loss1}.

\begin{table*}[!htbp]
\newcolumntype{Y}{>{\arraybackslash}X}
\small
\centering
\begin{tabularx}{\linewidth}{ll Y}
\toprule

$\texttt{Diff100}$ & \textbf{Question} & \emph{How many percent of people were not Hispanic?} \\
 & \textbf{Passage} &
The 2010 United States Census reported that Lassen County had a population of 34,895. The racial makeup of Lassen County was 25,532 (73.2\%) White (U.S. Census), 2,834 (8.1\%) African American (U.S. Census), 1,234 (3.5\%) Native American (U.S. Census), 356 (1.0\%) Asian (U.S. Census), 165 (0.5\%) Pacific Islander (U.S. Census), 3,562 (10.2\%) from Race (United States Census), and 1,212 (3.5\%) from two or more races.  Hispanic (U.S. Census) or Latino (U.S. Census) of any race were 6,117 persons (17.5\%). \\
\cmidrule(lr){2-3}
& \textbf{Answer} & 82.5\\
& \textbf{Prediction} & $\texttt{Diff100}(17.5) = 100 - 17.5 = 82.5$ \\

\cmidrule{2-3}

& \textbf{Question} & 
\emph{How many percent were not from 18 to 24?} \\
& \textbf{Passage} &
In the city, the population was spread out with 12.0\% under the age of 18, 55.2\% from 18 to 24, 15.3\% from 25 to 44, 10.3\% from 45 to 64, and 7.1\% who were 65 years of age or older.  The median age was 22 years. For every 100 females, there were 160.7 males.  For every 100 females age 18 and over, there were 173.2 males. \\
\cmidrule(lr){2-3}
& \textbf{Answer} &
44.8\\
& \textbf{Prediction} &
$\texttt{Diff100}(55.2) = 100 - 55.2 = 44.8$ \\

\midrule

$\texttt{Sum}$ & \textbf{Question} & 
\emph{How many millions of people in all did Germany and France have as residents that were born outside the EU?} \\
& \textbf{Passage} &
In 2010, 47.3 million people who lived in the EU were born outside their resident country. This corresponds to 9.4\% of the total EU population. Of these, 31.4 million (6.3\%) were born outside the EU and 16.0 million (3.2\%) were born in another EU member state. The largest absolute numbers of people born outside the EU were in Germany (6.4 million), France (5.1 million), the United Kingdom (4.7 million), Spain (4.1 million), Italy (3.2 million), and the Netherlands (1.4 million). \\
\cmidrule(lr){2-3}
& \textbf{Answer} &
11.5 \\
& \textbf{Prediction} &
$\texttt{Sum}(6.4, 5.1)  = 6.4 + 5.1 = 11.5$ \\

\midrule

$\texttt{Sum3}$ & \textbf{Question} & 
\emph{How many people, households, and families reside in the county according to the 2000 census?} \\
& \textbf{Passage} & 
As of the census of 2000, there were 40,543 people, 15,416 households, and 11,068 families residing in the county. The population density was 99 people per square mile 
(38/km${}^\text{2}$). There were 16,577 housing units at an average density of 40 per square mile 
(16/km${}^\text{2}$). The racial makeup of the county was 95.99\% Race (United States Census), 2.19\% Race (United States Census) or Race (United States Census), 0.26\% Race (United States Census), 0.38\% Race (United States Census), 0.20\% from Race (United States Census), and 0.97\% from two or more races. 0.66\% of the population were Race (United States Census) or Race (United States Census) of any race. 29.3\% were of united states, 22.2\% germans, 12.1\% english people and 10.9\% irish people ancestry according to 2000 United States Census. \\
\cmidrule(lr){2-3}
& \textbf{Answer} &
67027 \\
& \textbf{Prediction} &
$\texttt{Sum3}(\texttt{Sum}(40543,15416),11068) = 40543 + 15416 + 11068 = 67027$ \\

\midrule

$\texttt{Diff}$ & \textbf{Question} & 
\emph{How many years after the king's brothers raised a rebellion did Ava cede all northern Avan territory down to present-day Shwebo to Mohnyin?}\\
& \textbf{Passage} & 
Ava's authority deteriorated further in Shwenankyawshin's reign . Three of the king's own brothers openly raised a rebellion in 1501. Mohnyin, Ava's former vassal, now began to raid its territory. In 1507, Ava ceded to Mohnyin all northern Avan territory down to present-day Shwebo in the vain hope that the raids would stop. It did not. Ava desperately tried to retain Toungoo's loyalty by ceding the key Kyaukse granary to Toungoo but it too failed. Toungoo took the region but formally broke away in 1510. Ava's only ally was the Shan state of Thibaw , which too was fighting Mohnyin's raids on its territory. Mohnyin was attacking other Shan states when it was not raiding Ava. It seized Bhamo from Thibaw in 1512 in the east, and raiding Kale in the west. The Ava-Thibaw alliance was able to retake Shwebo for a time but Mohnyin proved too strong. By the early 1520s, Chief Sawlon of Mohnyin had assembled a confederation of Shan states  under his leadership. Prome had also joined the confederation. The confederation wiped out Ava's defences in Shwebo in 1524. Finally on 25 March 1527, the forces of the confederation and Prome took Ava. The Confederation later sacked Prome in 1533 because Sawlon felt that Prome had not given sufficient help. \\
\cmidrule(lr){2-3}
& \textbf{Answer} & 6 \\
& \textbf{Prediction} &
$\texttt{Diff}(1501,1507) = 1507 - 1501 = 6$ \\

\bottomrule
\end{tabularx}

\caption{Correct predictions of our model. Examples are from the DROP development set.}
\label{tab:win1}
\end{table*}

\begin{table*}[!htbp]
\newcolumntype{Y}{>{\arraybackslash}X}
\small
\centering
\begin{tabularx}{\linewidth}{ll Y}
\toprule

$\texttt{Div}$ &  \textbf{Question} &
Eric has 9306 erasers. If he shares them among 99 friends, how many erasers does each friend get? \\
\cmidrule(lr){2-3}
& \textbf{Answer} & 94.0 \\
& \textbf{Prediction} & 
$\texttt{Div}(9306,99) = 9306 / 99 = 94$ \\

\midrule

$\texttt{Mul}$ & \textbf{Question} &
It took Katherine 3 hours to run to Louis's house at 8 miles per hour. How far is it between Katherine's house and Louis's house? \\
\cmidrule(lr){2-3}
& \textbf{Answer} & 24  \\
& \textbf{Prediction} & $\texttt{Mul}(3, 8) = 3 * 8 = 24$ \\

\midrule

$\texttt{Merge}$ & \textbf{Question} &
What professionals paid higher rates than the advocates? \\
& \textbf{Passage} &
The poll tax was resurrected during the 17th century, usually related to a military emergency. It was imposed by Charles I of England in 1641 to finance the raising of the army against the Scottish and Irish uprisings. With the Restoration (England) of Charles II of England in 1660, the Convention Parliament (1660) instituted a poll tax to finance the disbanding of the New Model Army (pay arrears, etc.) (12 Charles II c.9). The poll tax was assessed according to "rank", e.g. dukes paid £100, earls £60, knights £20, esquires £10. Eldest sons paid 2/3rds of their fathers rank, widows paid a third of their late husbands rank. The members of the livery companies paid according to companys rank (e.g. masters of first-tier guilds like the Mercers paid £10, whereas masters of fifth-tier guilds, like the Clerks, paid 5 shillings). Professionals also paid differing rates, e.g. physicians (£10), judges (£20), advocates (£5), attorneys (£3), and so on. Anyone with property (land, etc.) paid 40 shillings per £100 earned, anyone over the age of 16 and unmarried paid 12-pence and everyone else over 16 paid 6-pence. \\
\cmidrule(lr){2-3}
& \textbf{Answer} & ``physicians'', ``judges'' \\
& \textbf{Prediction} & $\texttt{Merge}(\text{physicians}, \text{judges}) = \text{``physicians'', ``judges''} $ \\

\bottomrule
\end{tabularx}

\caption{Correct predictions from our model. Examples are from the DROP and IL development sets.}
\label{tab:win2}
\end{table*}

\begin{table*}[!htbp]
\newcolumntype{Y}{>{\arraybackslash}X}
\newcolumntype{L}{>{\hsize=.15\hsize}Y}
\newcolumntype{R}{>{\hsize=.85\hsize}Y}
\small
\centering
\begin{tabularx}{\linewidth}{L l R}
\toprule

Counting error &  \textbf{Question} &
\emph{How many touchdown passes were there during the second half?} \\
& \textbf{Passage} &
Coming off their home win over the Lions, the 49ers flew to the Louisiana Superdome for a Week 4 duel with the New Orleans Saints.  In the first quarter, the Niners struck first as kicker Joe Nedney got a 47-yard field goal.  In the second quarter, the Saints took the lead with QB Drew Brees completing a 5-yard and a 33-yard TD pass to WR Lance Moore.  San Francisco would answer with Nedney's 49-yard field goal, yet New Orleans replied with Brees' 47-yard TD pass to WR Robert Meachem. In the third quarter, the 49ers tried to rally as Nedney kicked a 38-yard field goal.  However, in the fourth quarter, the Saints continued to pull away as RB Deuce McAllister got a 1-yard TD run.  The Niners tried to rally as QB J.T. O'Sullivan completed a 5-yard TD pass to WR Isaac Bruce, yet New Orleans sealed the win with kicker  Martín Gramática  nailing a 31-yard field goal. \\
\cmidrule(lr){2-3}
& \textbf{Answer} & 1 ($\texttt{Literal}$)\\
& \textbf{Prediction} &  3 ($\texttt{Literal}$)\\

\midrule

Correct for the wrong reason &  \textbf{Question} &
\emph{How many yards was the shortest touchdown run?} \\
& \textbf{Passage} &
The Steelers went back home for another showdown with the Patriots. This game is notable for being the very first game that QB Ben Roethlisberger would miss out on against the Patriots. In the first quarter, The Patriots scored first when Tom Brady found James White on a 19-yard touchdown pass for a 7-0 lead for the only score of the period. In the second quarter, they increased their lead when LaGarrette Blount ran for a 3-yard touchdown to make it 14-0. The Steelers got on the board later on in the quarter when Landry Jones found Darrius Heyward-Bey on a 14-yard touchdown pass for a 14-7 game. The Steelers closed out the scoring of the first half when Chris Boswell kicked a 32-yard field goal for a 14-10 game at halftime. In the third quarter, the Steelers went back to work as Boswell kicked another field goal to get his team within 1, 14-13 from 46 yards out. The Pats pulled away later on when Brady found Rob Gronkowski on a 36-yard touchdown pass (with a failed PAT) for a 20-13 game. In the fourth quarter, the Steelers came within 4 again when Boswell made a 44-yard field goal for a 20-16 game. But the Pats sealed the game after Blount ran for a 5-yard touchdown and the eventual final score of 27-16. With the loss, the Steelers went into their bye week at 4-3. Regardless, due to the Ravens' loss to the Jets, they still remain in first place in the AFC North. The team dropped to 0-1 on the season without Roethlisberger as a starter and their seven-game home winning streak was snapped. \\
\cmidrule(lr){2-3}
& \textbf{Answer} & ``3'' ($\texttt{Span}$)\\
& \textbf{Prediction} & $\texttt{Sum}(3, 0) = 3 $  \\

\midrule

Wrong type & \textbf{Question} &
\emph{How many points did the Lions score in the first half?} \\
& \textbf{Passage} &
For their annual Thanksgiving Day game, the Lions hosted a rematch with their divisional rival, the Minnesota Vikings. The Vikings scored 13 points in the first quarter via a one-yard touchdown pass from Case Keenum to Kyle Rudolph, and a nine-yard touchdown run from Keenum. The Lions responded with 10 points in the second quarter via a 32-yard field goal from Matt Prater and a six-yard touchdown pass from Matthew Stafford to Marvin Jones Jr. The Vikings extended their lead in the second quarter via a 22-yard touchdown pass from Keenum to Rudolph to make the score 20-10 in favor of Minnesota at half-time. The Vikings opened the scoring in the second half via a two-yard touchdown run from Latavius Murray. The Lions responded with two field goals from Prater in the third quarter from 32-yards, and 50-yards, respectively. The Lions reduced the Vikings lead to four points in the fourth via a 43-yard touchdown pass from Stafford to Jones. The Vikings extended their lead in the fourth quarter via a 36-yard field goal from Kai Forbath. The Lions' attempted comeback failed when Stafford's pass intended for Jones was intercepted by Xavier Rhodes. On the Vikings' ensuing drive, Forbath's 25-yard field goal attempt was blocked by Darius Slay and recovered by Nevin Lawson and returned for a 77-yard touchdown, which was then nullified due to an offside penalty on Slay, making the final score 30-23 in favor of Minnesota, snapping the Lions' three-game winning streak. \\
\cmidrule(lr){2-3}
& \textbf{Answer} &  ``10'' ($\texttt{Span}$) \\
& \textbf{Prediction} & 7 ($\texttt{Literal}$) \\

\bottomrule
\end{tabularx}

\caption{Examples of wrong predictions from our model on the DROP development set.}
\label{tab:loss1}
\end{table*}

\bibliography{emnlp-ijcnlp-2019}
\bibliographystyle{acl_natbib}